\documentclass[final]{l4dc2026}

\title[Deep QP Safety Filter]{Deep QP Safety Filter: \\ Model-free Learning for Reachability-based Safety Filter}
\usepackage{times}
\usepackage{kotex}
\usepackage{pifont}
\usepackage{amsfonts}
\usepackage{mathtools}
\usepackage{array}
\usepackage{textcomp}
\usepackage{stfloats}
\usepackage{verbatim}
\usepackage{booktabs} 
\usepackage{xcolor}
\usepackage{makecell}
\usepackage{multirow} 
\usepackage{siunitx} 
\usepackage{ragged2e}
\usepackage{bm} 

\usepackage{algorithm}
\usepackage{algpseudocode}
\usepackage{graphicx}
\usepackage{caption}

\usepackage{thmtools}

\declaretheorem{theorem}
\declaretheorem{lemma}
\declaretheorem{definition}
\declaretheorem{corollary}
\declaretheorem[style=remark]{remark} %

\usepackage[compact]{titlesec}

\titlespacing*{\section}{0pt}{0.8\baselineskip}{0.3\baselineskip}
\titlespacing*{\subsection}{0pt}{0.7\baselineskip}{0.3\baselineskip}
\titlespacing*{\subsubsection}{0pt}{0.6\baselineskip}{0.3\baselineskip}

\author{%
 \Name{Byeongjun Kim} \Email{qudwns3456@snu.ac.kr}\\
 \addr Seoul National University
 \AND
 \Name{H. Jin Kim} \Email{hjinkim@snu.ac.kr}\\
 \addr Seoul National University
}

\begin{document}

\maketitle
\vspace{-20pt}
\begin{abstract}%
We introduce Deep QP Safety Filter, a fully data-driven safety layer for black-box dynamical systems.
Our method learns a Quadratic-Program (QP) safety filter without model knowledge by combining Hamilton–Jacobi (HJ) reachability with model-free learning. We construct contraction-based losses for both the safety value and its derivatives, and train two neural networks accordingly. In the exact setting, the learned critic converges to the viscosity solution (and its derivative), even for non-smooth values.
Across diverse dynamical systems -- even including a hybrid system -- and multiple RL tasks, Deep QP Safety Filter substantially reduces pre-convergence failures while accelerating learning toward higher returns than strong baselines, offering a principled and practical route to safe, model-free control.

\end{abstract}


\section{Introduction}
Safety-critical control has advanced rapidly in recent years, revealing impressive capabilities across robotics and autonomous systems. Among these, filtering-based approaches that modify an unsafe reference input into a safe one have become popular for their simplicity and real-time feasibility~\citep{hsu2023safety_survey}. Within this family, Quadratic-Program (QP) formulations that enforce safety even inside the safe set often produce smoother behavior than switching-based methods, and have been deployed on challenging, unstable platforms such as bipedal robots~\citep{clf_cbf_qp_bipdeal}.

Despite this progress, current filters depend heavily on prior model knowledge: their reliability deteriorates as model uncertainty grows. Recent learning-based variants attempt to mitigate uncertainty, yet many still presume partial access to the dynamics, which is unavailable for black-box systems. This creates a practical gap between elegant theory and deployment on real platforms with unknown or changing dynamics.

We address this gap by proposing a framework that learns a QP-based safety filter directly from data, requiring no model knowledge. Our method integrates time-discounted Hamilton–Jacobi (HJ) reachability with a model-free learning paradigm to construct a filter that preserves the desirable control characteristics of QP methods while enforcing data-driven safety constraints. Theoretical guarantees follow from the contraction properties of Bellman operators, yielding convergence and enabling a practical, model-free implementation.

We validate the approach in extensive simulations. The learned critic converges to the analytic HJ solution, and, when used as a safety filter, significantly reduces training failures and improves RL policy performance relative to baselines. The filter’s conservatism is reduced via HJ reachability and can be tuned in aggressiveness by the user, allowing reuse across tasks. Finally, we further test the method on a hybrid system that does not meet the theoretical assumptions. Its strong performance indicates that the assumptions are practically valid when approximated by neural networks.

\begin{figure}[t]
    \centering
    \includegraphics[width=0.95\linewidth]{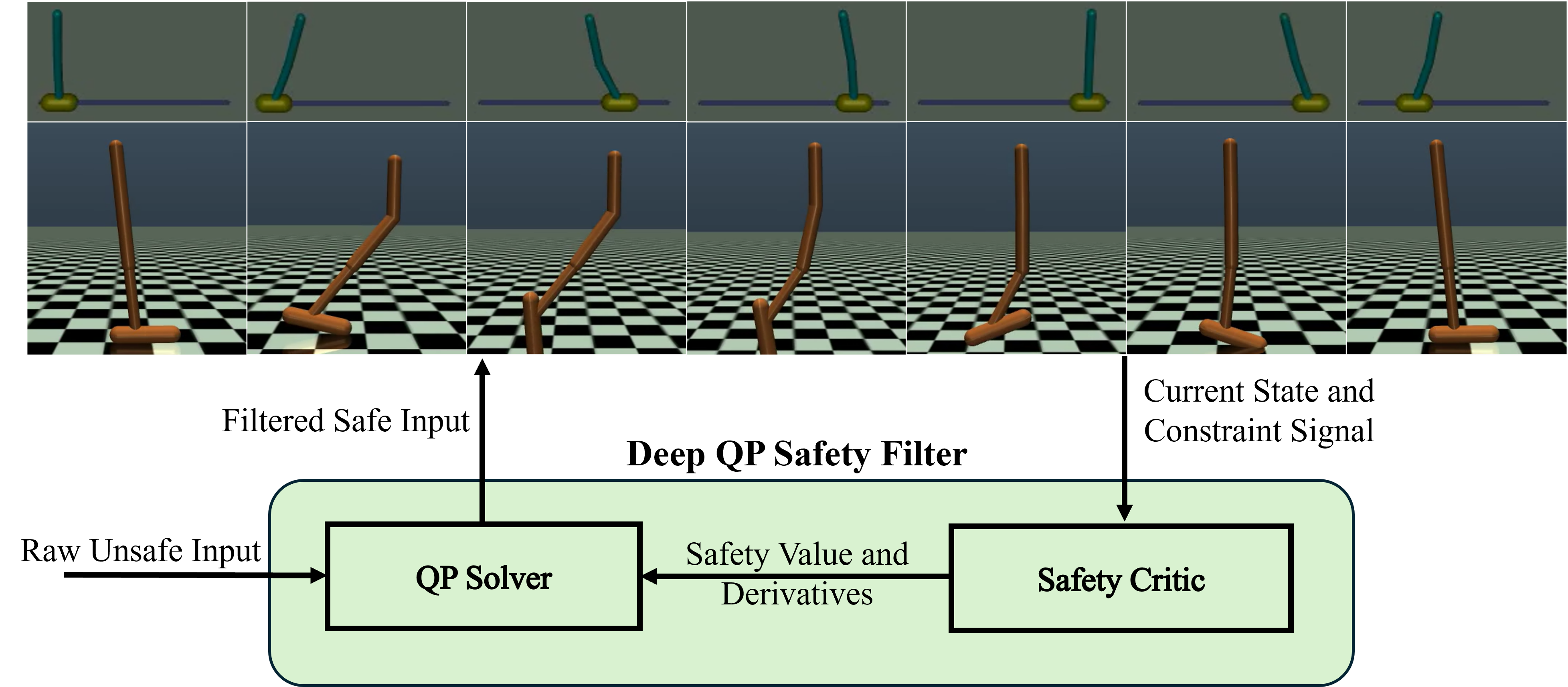}
        \caption{Overview of the Deep QP Safety Filter learned purely from transition data. The Safety Critic maps state and constraints to a safety value and derivatives, used by a QP solver together with a raw reference input, in order to produce a safe control input. The example sequence illustrates an unstable system with bang-bang ($\pm 1$) reference commands, where the filtered commands under our method maintain safety.}
        \label{fig: thumnail}
\vspace{-15pt}
\end{figure}

\section{Related Works}
\textbf{Learning with model assumptions.} A large body of research uses learning to overcome the limitations of model-based methods, such as the curse of dimensionality in HJ reachability~\citep{bansal2021deepreach,hsu2023isaacs,JingqiLiTimediscounted} or the difficulty of handcrafting Control Barrier Functions (CBFs)~\citep{robey2020learning,xiao2023barriernet,zhang2025gcbf+}. Other works combine CBF-QPs with RL to compensate for imperfect nominal models~\citep{gangopadhyay2022safe,choi2020reinforcement}. However, these approaches still rely on partial model knowledge, such as a pre-defined CBF structure~\citep{cbf_learning_model_mismatchovercome}, properties like Lipschitz constants~\citep{choi2025_online_reachability_compute}, or a precomputed safe policy~\citep{lemma1_shortened} -- assumptions that are rarely available in the black-box settings. \\
\textbf{Model-free safety learning.} Inspired by RL, model-free safety learning has also been widely explored. 
Many of these methods rely on discounted cumulative costs~\citep{gu2024review_discounted_sum_of_cost_safeRL}, which do not ensure persistent satisfaction of state constraints. 
Others aim for model-free state constraint satisfaction but may incur numerous safety violations during training~\citep{yu2022reachability} or produce switching filters~\citep{fisac2019bridging}. 
This switching behavior is unsuitable for highly dynamic systems, where such rapidly changing safe control strategies are not preferred.
More recent work learns discriminating hyperplanes in the input space~\citep{lavanakul2024_discriminatinghyperplane}. 
However, for black-box systems without prior safety knowledge, one must adopt the RL-based training approach of this method. Since this RL approach simply maximizes survival time, the absence of reachability foundations may lead to conservative behavior.

\section{Method}
In this section, we describe the overall setting of our problem and introduce the HJ reachability as a preliminary requirement for the proposed method. We then propose a model-free safety filter learning framework grounded on the HJ reachability.

\subsection{Problem Setup}
Let us denote the state and the control input as $x \in \mathcal{X} \subseteq \mathbb{R}^n$ and $u\in \mathcal{U}\subset \mathbb{R}^m$, respectively, where $\mathcal{X}$ and $\mathcal{U}$ represent the state space and the input space.
Since most actuators are subject to hardware limits, we consider the input space $\mathcal{U}$ to be constrained by a known pair of $(A,b)$, i.e., 
{\setlength\abovedisplayskip{4pt}%
 \setlength\belowdisplayskip{4pt}%
\[
\mathcal{U}=\{u\in\mathbb{R}^m \;\vert\; Au\leq b\}.
\]}
This also includes the common box-constrained case in which each actuator independently possesses its own upper and lower bounds.

The system provides a differentiable output signal $c(x) \in \mathbb{R}$ whose functional form is unknown, although its value can be directly measured.
We impose the state constraint $c(x) \geq 0$. This constraint can be viewed as an observable signal analogous to a reward signal in RL. Observations are made at a known fixed time interval $\delta t > 0$. Since sensing and control are performed discretely in real systems, we use transition data sampled at the same fixed interval $\delta t$. That is, given a particular state $x$, applied control input $u$, observed constraint signal $c$, and the next state and constraint $x_{\delta t}^u$ and $c_{\delta t}^u$ after $\delta t$, we form transitions of the type $(x,u,c,x^u_{\delta t}, c^u_{\delta t})$.

We assume the control-affine dynamics of the given black-box system:
{\setlength\abovedisplayskip{4pt}%
 \setlength\belowdisplayskip{4pt}%
\[
\dot{x} = f(x) + g(x)u.
\]}
This assumption is well justified, as many physical systems are controlled by force or torque commands through actuators.
In such cases, the Euler–Lagrange equation naturally leads to a control-affine form.
Even if a system is not strictly control-affine, the structure can still be represented by introducing a virtual input $v$ with $\dot{u}=v$.
Moreover, as will be discussed later, even a highly dynamic hybrid system exhibiting discontinuous state transitions can be effectively learned in a smooth manner through discrete transition data.
The proposed algorithm has been empirically verified to handle such hybrid cases successfully.

Lastly, we define the notion of safety considered throughout this work.
{\setlength\abovedisplayskip{4pt}%
 \setlength\belowdisplayskip{4pt}%
\begin{definition}[Safe state and safe set]
    A state $x$ is said to be safe if there exists a control trajectory such that the state at time $t$, $\bm{x}(t)$, driven by that control from initial condition $\bm{x}(0) = x$, satisfies $c(\bm{x}(t)) \geq 0$ for all $t\geq0$.
    A set $S$ is said to be a safe set if every element $x\in S$ is a safe state.
\end{definition}}

Our goal is to design a QP-based safety filter for systems with black-box dynamics.
Unlike bang–bang safety filters that intervene only when the safety value falls below a threshold, we aim for a formulation that consistently filters the control input.
This approach provides smoother control commands and more stable system behavior while preserving safety guarantees.
\subsection{Hamilton-Jacobi Reachability}
Hamilton--Jacobi (HJ) reachability has been one of the cornerstones in the field of safety-critical control.
It is an optimal-control-based framework that handles input bounds and relative degrees between constraint and control, enabling safe control with minimal conservatism.
Let us consider the safety value function and its corresponding Partial Differential Equation (PDE) \citep{fisac2019bridging, fisac2015reach}: 
{\setlength\abovedisplayskip{4pt}
 \setlength\belowdisplayskip{4pt}
\begin{equation*}
    V(x)= \sup_{\pi} \inf_{t\geq0} c(x^\pi_t ), \quad \min \left\{c(x) - V(x), \max_{u\in\mathcal{U}} \frac{\partial V}{\partial x} (x)(f(x)+g(x)u) \right\} = 0.
\end{equation*}}
where $\pi$ denotes any control policy and $x_t^{\pi}$ is the state at time $t$ driven by policy $\pi$ starting from $x$.
The function $V(x)$ represents the smallest achievable constraint value when the system is controlled optimally from the current state $x$.
By definition, the current state $x$ is considered \emph{safe} if and only if $V(x)\ge0$, and the $0$-superlevel set of $V(x)$ renders the largest control invariant safe set.

To enable model-free learning, we rewrite the definition of $V(x)$ in a recursive form:
{\setlength\abovedisplayskip{4pt}%
 \setlength\belowdisplayskip{4pt}%
\begin{equation} \label{original_value_recursive_form}
    V(x) = \sup_\pi \min \left\{ \min_{0\leq t_1\leq s} c(x^
    \pi _{t_1}), \inf_{t_2 \geq s} c(x^\pi_{t_2}) \right\} = \sup_\pi \min \left\{ \min_{0\leq t\leq s} c(x^\pi _{t}), V(x^\pi_{s}) \right\}
\end{equation}}
However, \eqref{original_value_recursive_form} requires transition data to be collected under an optimally safe policy. Since such a policy is not known in advance, this formulation is not directly suitable for model-free learning.
To relax this limitation and allow learning from arbitrary transition data, we introduce a state-action safety value function analogous to the $Q$-function in RL \citep{rl_Q_learning}:
{\setlength\abovedisplayskip{4pt}%
 \setlength\belowdisplayskip{4pt}%
\begin{equation} \label{original_Q_recursive_form}
    Q_s(x,u) = \min \left\{ \min_{0\leq t_1\leq s} c(x^
    u _{t_1}), \sup_\pi \inf_{t_2 \geq 0} c((x^
    u _{s})^\pi_{t_2}) \right\} =\min \left\{ \min_{0\leq t\leq s} c(x^u _{t}), V(x^u_{s}) \right\}
\end{equation}}
Here, $Q_s(x,u)$ represents the smallest constraint value that can be achieved when applying a constant control $u$ for a short horizon $s\ge0$ and then switching to the optimally safe policy thereafter.
This formulation serves as the foundation for a model-free learning algorithm that can learn directly from arbitrary transition data.
\subsection{Time-discounted reachability}
The recursive definitions in \eqref{original_value_recursive_form} and \eqref{original_Q_recursive_form} do not inherently possess contraction properties, which makes them unsuitable for stable critic learning. 
To overcome this limitation, several works \citep{fisac2019bridging, JingqiLiTimediscounted} introduced a discount factor in the discrete-time domain.
In this section, we extend this concept to the continuous-time setting and establish the theoretical foundations required to learn the corresponding discounted value functions and their derivatives.
First, for a non-negative scalar $\lambda \geq 0 $, we define the discounted value functions $V^\lambda (x)$ and $Q^\lambda _s(x,u)$ as
{\setlength\abovedisplayskip{4pt}
 \setlength\belowdisplayskip{4pt}
\begin{align} 
V^\lambda (x) := & \sup_{\pi} \inf_{t\geq0} \left\{ \int_{0}^{t}{\lambda e^{-\lambda \tau} c(x^\pi_\tau ) d\tau} + e^{-\lambda t} c(x^\pi_t ) \right\}, \label{discounted V} \\ 
Q^\lambda _s (x,u) := & \min \left\{ \min_{0\leq t\leq s} \int_{0}^{t}{\lambda e^{-\lambda \tau} c(x^u_\tau ) d\tau} + e^{-\lambda t} c(x^u_t ), \int_{0}^{s}{\lambda e^{-\lambda \tau} c(x^u_\tau ) d\tau}+e^{-\lambda s}V^\lambda(x^u_{s}) \right\} \label{discounted Q}
\end{align}}
We list several corollaries relating the original and discounted value functions.
{\setlength\abovedisplayskip{4pt}
 \setlength\belowdisplayskip{4pt}
\begin{corollary} \label{corollary: CVQ order}
    For all $(x,u)\in\mathcal{X}\times\mathcal{U}$, the following inequalities hold:
\[
Q_s(x,u) \le V(x) \le c(x),
\qquad
Q_s^\lambda(x,u) \le V^\lambda(x) \le c(x).
\]
\end{corollary}
\begin{corollary} \label{corollary: discounting approach}
    As the discount factor approaches zero, the discounted functions converge to their undiscounted counterparts:
\[
\lim_{\lambda\downarrow 0} V^\lambda(x) = V(x), 
\qquad 
\lim_{\lambda\downarrow 0} Q_s^\lambda(x,u) = Q_s(x,u),
\quad \forall (x,u)\in\mathcal{X}\times\mathcal{U}.
\]
\end{corollary}
\begin{corollary} \label{corollary: Q goes to V as s goes to 0}
    As the duration $s$ tends to zero, the state-action safety values converge to their corresponding safety values:
\[
\lim_{s\downarrow 0} Q_s(x,u) = V(x), 
\qquad 
\lim_{s\downarrow 0} Q_s^\lambda(x,u) = V^\lambda(x), \quad \forall (x,u)\in\mathcal{X}\times\mathcal{U}.
\]
\end{corollary}}
\begin{proof}
    It directly follows from the definitions in \eqref{original_value_recursive_form} and \eqref{original_Q_recursive_form} that $V(x)\leq c(x)$ and $V^\lambda(x)\leq c(x)$ for all $x\in\mathcal{X}$. 
    Because $V^\lambda(x)$ is defined as the supremum over all admissible control , $Q_s(x,u)\leq V(x)$ and $Q_s^\lambda(x,u)\leq V^\lambda(x)$ for all $(x,u)\in\mathcal{X} \times \mathcal{U}$. These prove the first corollary.
    The second and third corollaries are obtained by substituting $\lambda=0$ and $s=0$ into \eqref{discounted V}-\eqref{discounted Q} and \eqref{original_Q_recursive_form}-\eqref{discounted Q}.
\end{proof}
These results serve as the starting point for constructing our model-free safety filter learning algorithm. For further analysis, we consider the following HJ PDE for $V^\lambda(x)$.

\begin{lemma}[PDE for the discounted safety value function] \label{lemma: discounted V PDE}
    The value function in \eqref{discounted V} satisfies the following PDE in the viscosity sense \citep{lemma1_shortened},\citep{lemma1_origin}:
    \begin{equation} \label{eq: discounted V PDE}
        \min \left\{ c(x) - V^\lambda(x), \max_{u\in\mathcal{U}} \frac{\partial V}{\partial x} (x)(f(x)+g(x)u) + \lambda \left( c(x)-V^\lambda(x)\right) \right\} =0
    \end{equation}
\end{lemma}
The PDE in \textbf{Lemma \ref{lemma: discounted V PDE}} characterizes the local optimality condition of the discounted safety value function.
To make this relation explicit in terms of the control input $u$, we next introduce an auxiliary quantity that measures the instantaneous change of $V^\lambda(x)$ along the system dynamics. This quantity, referred to as the advantage of discounted safety value, plays a key role in defining the model-free Bellman operator used in our learning framework. To facilitate further analysis, we assume that $V^\lambda(x)$ is differentiable. This assumption not only simplifies the subsequent derivations but is also not restrictive in practice, as most neural network approximators employ smooth activation functions such as \texttt{tanh} or the Exponential Linear Unit (ELU)~\citep{elu_activation}. 

\begin{theorem}[Advantage in discounted safety value] \label{theorem: small q lambda}
    Let us define the advantage $q^\lambda(x,u) :=\\ \lim_{s\downarrow0} \frac{Q^\lambda _s (x,u) - V^\lambda(x)}{s}$. Then, $q^\lambda$ satisfies the following:
    {\setlength{\jot}{0pt}
    \begin{gather*}
        q^\lambda(x,u) \leq0 \; \forall (x,u)\in\mathcal{X}\times\mathcal{U}, \qquad \max_{u\in\mathcal{U}} q^\lambda(x,u) = 0 \; \forall x\in\mathcal{X}, \\
        q^\lambda(x,u) = \min \left\{ c(x) - V^\lambda(x), \frac{\partial V^\lambda}{\partial x} (x)(f(x)+g(x)u) + \lambda \left( c(x)-V^\lambda(x)\right) \right\}.
    \end{gather*}}
\end{theorem}
\begin{proof}
    Expanding each term inside the $\min$ of \eqref{discounted Q} with a sufficiently small $s$,
{\setlength{\jot}{0pt}
\setlength\abovedisplayskip{4pt}
 \setlength\belowdisplayskip{4pt}
    \begin{align*}
        \min_{0\leq t\leq s} \int_{0}^{t}{\lambda e^{-\lambda \tau} c(x^u_\tau ) d\tau} + e^{-\lambda t} c(x^u_t ) &= c(x) + s \min \left\{0, \frac{\partial c}{\partial x}(x)\left(f(x)+g(x)u\right) \right\} + O(s^2) \\
        \begin{split}
        \int_{0}^{s}{\lambda e^{-\lambda \tau} c(x^u_\tau ) d\tau}+e^{-\lambda s}V^\lambda(x^u_{s}) &= V^\lambda(x)+ s \lambda \left(c(x) - V^\lambda(x)\right) 
        \\ & \quad + s \frac{\partial V^\lambda}{\partial x}(x)\left(f(x)+g(x)u\right) + O(s^2). 
        \end{split}
    \end{align*}}
    Now, we subtract $V^\lambda(x)$, divide $s$, and take $s \downarrow 0$. 
    For the ease of further analysis, we separately consider the two cases: 1) $V^\lambda(x)=c(x)$ and 2) $V^\lambda(x) < c(x)$. 
    For the case of 1), we have the following:
    \begin{equation*}
        q^\lambda(x,u) = \min \left\{ 0, \frac{\partial V^\lambda}{\partial x}(x)\left(f(x)+g(x)u\right) \right\},
    \end{equation*}
    where we used the fact that, from \textbf{Corollary~\ref{corollary: CVQ order}} and the continuity of both $c(x)$ and $V^\lambda(x)$, \[ \frac{\partial V^\lambda}{\partial x}(x)\left(f(x)+g(x)u\right)     \leq  \frac{\partial c}{\partial x}(x)\left(f(x)+g(x)u\right) \; \forall u\in\mathcal{U} \quad \text{if} \quad V^\lambda(x)=c(x).\]
    When $V^\lambda(x)<c(x)$, the $\min$ operator activates the second expression. 
    In this case, we have $q^\lambda(x,u) = \frac{\partial V^\lambda}{\partial x}(x)\left(f(x)+g(x)u\right) + \lambda \left(c(x) - V^\lambda(x)\right)$.
    Therefore, by the \textbf{Corollary \ref{corollary: CVQ order}} and \textbf{Lemma \ref{lemma: discounted V PDE}}, we can concisely write $q^\lambda(x,u)$ as the following:
    \begin{equation*}
        q^\lambda(x,u)=\min \left\{ c(x) - V^\lambda(x), \frac{\partial V^\lambda}{\partial x} (x)(f(x)+g(x)u) + \lambda \left( c(x)-V^\lambda(x)\right) \right\}.
    \end{equation*}
    Thus, $\max_{u\in\mathcal{U}} q^\lambda(x,u)=0 \; \forall x\in\mathcal{X}$ by \textbf{Lemma \ref{lemma: discounted V PDE}}, which means $q^\lambda(x,u)\leq0 \; \forall (x,u)\in\mathcal{X}\times \mathcal{U}$.
\end{proof}
\vspace{-5pt}
\subsection{Model-free safety filter learning}
In this subsection, we construct a model-free safety filter using the time-discounted reachability formulation developed in the previous subsection. 
For a sampled transition $(x, u, c, x^u_{\delta t}, c^u_{\delta t})$ observed at a fixed interval $\delta t$,  the first term in \eqref{discounted Q} can be treated as monotonic over $0 \le t \le \delta t$. 
By leveraging \textbf{Corollary \ref{corollary: CVQ order}} and \textbf{Theorem \ref{theorem: small q lambda}}, we can rewrite \eqref{discounted Q} as the following:
{\setlength\abovedisplayskip{4pt}
 \setlength\belowdisplayskip{4pt}
\begin{align} \label{value bellman equation}
    Q^\lambda_{\delta t}(x,u) \simeq V^\lambda(x) + \delta t \cdot q^\lambda(x,u) \simeq \min \left\{ c, \texttt{int}(c,\lambda,\delta t) + e^{-\lambda \cdot \delta t}V^\lambda (x^u_{\delta t}) \right\}
\end{align}}
where, \texttt{int}$(c,\lambda, \delta t)$ denotes a numerically approximiated integral term $\int_{0}^{\delta t} \lambda e^{-\lambda \tau }c(x^u_\tau) d\tau$.
From \textbf{Theorem \ref{theorem: small q lambda}}, $\max_{u\in\mathcal{U}} q^\lambda(x,u)=0$ for all $x\in\mathcal{X}$.
Using this fact and $\frac{\partial V^\lambda}{\partial x}(x)(f(x)+g(x)u) \!\simeq\! (V^\lambda(x^u_{\delta t}) - V^\lambda(x))/\delta t$, we can write
{\setlength{\jot}{0pt}
\setlength\abovedisplayskip{4pt}%
 \setlength\belowdisplayskip{4pt}%
\begin{align} \label{derivatives bellman equation}
    \frac{\partial V^\lambda}{\partial x}(x)(f(x)+g(x)u) & \simeq 
        \frac{1}{\delta t}\cdot \Biggl( \min \biggl\{ c^u_{\delta t} -V^\lambda(x), e^{-\lambda \cdot \delta t}V^\lambda(x^u_{\delta t}) - V^\lambda(x) + \\
    & \phantom{\simeq}
        \texttt{int}(c^u_{\delta t}, \lambda, \delta t )+\delta t \cdot e^{-\lambda \cdot \delta t}\max_{u^\prime \in \mathcal{U}}\frac{\partial V^\lambda}{\partial x}(x^u_{\delta t})(f(x^u_{\delta t})+g(x^u_{\delta t})u^\prime) \biggr\} \Biggr) \nonumber
\end{align}}
We denote $\partial V^\lambda(x,u) := \frac{\partial V^\lambda}{\partial x}(x)(f(x)+g(x)u)$ and define two Bellman operators:
{\setlength{\jot}{0pt}
\setlength\abovedisplayskip{3pt}%
 \setlength\belowdisplayskip{3pt}%
\begin{equation} \label{bellman V}
    \mathcal{T}_v[V^\lambda](x,u):= \min \left\{ c, \texttt{int}(c,\lambda,\delta t) + e^{-\lambda \cdot \delta t}V^\lambda (x^u_{\delta t}) \right\} - \delta t \cdot q^\lambda(x,u)
\end{equation}
\begin{align} \label{bellman DV}
    \mathcal{T}_{dv}[\partial V^\lambda](x,u) := \Biggl\{\min & \biggl\{ c^u_{\delta t}, \texttt{int}(c^u_{\delta t},\lambda,\delta t) + e^{-\lambda \cdot \delta t}V^\lambda (x^u_{\delta t}) +  \\
    & \phantom{\simeq}
        \delta t \cdot e^{ -\lambda \cdot \delta t} \cdot \max_{u^\prime \in \mathcal{U}} \partial V^\lambda (x^u_{\delta t}, u^\prime)\biggr\} - V^\lambda(x) \Biggr\}  / \delta t \nonumber 
\end{align}}
When $q^\lambda$ in \eqref{bellman V} or $V^\lambda$ in \eqref{bellman DV} is treated as a fixed term independent of the function to which the Bellman operator is applied, each operator becomes an $e^{-\lambda \cdot \delta t}$-contraction in the supremum norm. This leads to the following results.
\begin{theorem}[{$e^{-\lambda \cdot \delta t}$-contraction on the discounted safety value}]\label{theorem: value contraction}
    Let two different value functions be $V^\lambda_1, V^\lambda_2$. For a fixed $q^\lambda$, $\left\Vert \mathcal{T}_v [V^\lambda_1] - \mathcal{T}_v [V^\lambda_2] \right\Vert_\infty \leq e^{-\lambda \cdot \delta t}\left\Vert V^\lambda_1 - V^\lambda_2\right\Vert_\infty$.
\end{theorem}
\begin{theorem}[{$e^{-\lambda \cdot \delta t}$-contraction on the derivatives}] \label{theorem: derivatives contraction}
    Let two different pairs of the derivatives be $\partial V^\lambda_1, \partial V^\lambda_2$. For a fixed safety value $V^\lambda$, $\left\Vert \mathcal{T}_{dv} [\partial V^\lambda_1] - \mathcal{T}_{dv} [\partial V^\lambda_2]\right\Vert_\infty \leq e^{-\lambda \cdot \delta t}\left\Vert \partial V^\lambda_1 - \partial V^\lambda_2\right\Vert_\infty$.
\end{theorem} 
\begin{proof} 
    Since $q^\lambda(x,u)$ in \eqref{bellman V} for both $V^\lambda_1$ and $V^\lambda_2$ are fixed and identical, the effect of this term vanishes when $\mathcal{T}_v[V^\lambda _1](x,u) - \mathcal{T}_v[V^\lambda _2](x,u)$ for all pair of $(x,u)$. Using $\lvert \min\{a,b\} - \min\{a,c\} \rvert \leq \lvert b - c \rvert$, we have the following inequality:
    \begin{align*}
        \left\vert \mathcal{T}_v [V^\lambda_1(x,u)] - \mathcal{T}_v [V^\lambda_2(x,u)]\right\vert \leq e^{-\lambda \cdot \delta t}\lvert V^\lambda _1 (x^u_{\delta t}) -V^\lambda _2 (x^u_{\delta t}) \rvert.
    \end{align*}
    Taking the supremum norm on $x$ and $u  $ for both sides concludes the proof of \textbf{Theorem \ref{theorem: value contraction}}.
    
    Similarly, terms related to $V^\lambda$ are canceled out in $\mathcal{T}_{dv} [\partial V^\lambda_1](x,u) - \mathcal{T}_{dv} [\partial V^\lambda_2](x,u)$. Thus,
    \begin{align*}
        \left\vert \mathcal{T}_{dv} [\partial V^\lambda_1](x,u) - \mathcal{T}_{dv} [\partial V^\lambda_2](x,u) \right\vert & \leq e^{-\lambda \cdot \delta t}\left\vert \max_{u_1 \in \mathcal{U}} \partial V^\lambda _1 (x^u_{\delta t}, u_1) -\max_{u_2 \in \mathcal{U}}\partial V^\lambda _2 (x^u_{\delta t}, u_2) \right\vert  \\ 
        & \leq e^{-\lambda \cdot \delta t} \max_{u^\prime \in \mathcal{U}}\left\vert \partial V^\lambda _1 (x^u_{\delta t}, u^\prime) - \partial V^\lambda _2 (x^u_{\delta t}, u^\prime) \right\vert
    \end{align*}
    Taking the supremum norm on $x \text{ and } u$ for both sides concludes the proof of \textbf{Theorem \ref{theorem: derivatives contraction}}.
\end{proof} 
These theorems imply that, for each fixed $q^\lambda$ or $V^\lambda$, 
repeated application of the operators converges to the corresponding unique fixed points.
In practice, following the standard approach in off-policy RL~\citep{rl_paper_sac, rl_paper_ddpg}, we adopt slowly updated target networks so that the $q^\lambda$ and $V^\lambda$ terms can evolve gradually during training, and the learned value and derivative functions progressively converge toward better fixed points.
We approximate $V^\lambda\!\approx\! v^\lambda_\theta$ and $\partial V^\lambda\!\approx\!\partial v^\lambda_\phi$, with parameters $(\theta,\phi)$. We denote parameters of target networks as $\theta^\prime$ and $\phi^\prime$, which are used to compute $q^\lambda \!\approx\! q^\lambda_{\theta^\prime, \phi^\prime}$ in \eqref{bellman V} and $V^\lambda \!\approx\! v^\lambda_{\theta^\prime}$ in \eqref{bellman DV}. Given a mini-batch of $N$ transitions $(x_i,u_i,c_i,x^u_{\delta t,i},c^u_{\delta t,i})_{i=1}^N$, we minimize the following losses:
{\setlength{\jot}{0pt}
\setlength\abovedisplayskip{4pt}
 \setlength\belowdisplayskip{4pt}
\begin{equation} \label{loss: V}
    \mathcal{L}_v = \frac{1}{N}\sum_{i=1}^{N} \Biggl\{ v^\lambda_\theta (x_i) - \min \left\{ c_i, \texttt{int}(c_i,\lambda,\delta t) + e^{-\lambda \cdot \delta t}v^\lambda_{\theta^\prime} (x^u_{\delta t, i}) \right\} + \delta t \cdot q^\lambda_{\theta^\prime, \phi^\prime}(x_i,u_i) \Biggl\}^2 
\end{equation}
\begin{align} \label{loss: derivatives}
    \mathcal{L}_{\partial v} = \frac{1}{N}\sum_{i=1}^{N} \Biggl\{ \partial v^\lambda_\phi (x_i, u_i) -& \Biggl\{\min \biggl\{ c^u_{\delta t, i}, \texttt{int}(c^u_{\delta t, i},\lambda,\delta t) + e^{-\lambda \cdot \delta t}v^\lambda_{\theta^\prime} (x^u_{\delta t, i}) +  \\
    & \phantom{\simeq}
        \delta t \cdot e^{ -\lambda \cdot \delta t} \cdot \max_{u^\prime \in \mathcal{U}} \partial v^\lambda_{\phi^\prime} (x^u_{\delta t, i}, u^\prime)\biggr\} - v^\lambda_{\theta^\prime}(x_i) \Biggr\} / \delta t \nonumber \Biggl\}^2
\end{align}}
The terms related to $x^u_{\delta t,i}$ are computed using the target networks for numerical stability. We further reparameterize $\partial v^\lambda_\phi(x,u) = a^\lambda_\phi(x)u - \max_{u\in\mathcal{U}}a^\lambda_\phi(x)u + b^\lambda_\phi(x)$, so that $\max_{u\in\mathcal{U}}\partial v^\lambda_\phi(x,u)=b^\lambda_\phi(x)$. This representation allows the condition from \textbf{Lemma~\ref{lemma: discounted V PDE}} to be imposed directly through the term $b^\lambda_\phi(\cdot)$.
In the sampled mini-batch, we enforce this condition by setting 
$b^\lambda_{\phi'}(x^u_{\delta t, i})=-\lambda (c(x^u_{\delta t,i})-v^\lambda_{\theta'}(x^u_{\delta t,i}))$ whenever $v^\lambda_{\theta'}(x^u_{\delta t,i})<c(x^u_{\delta t,i})$.

For data collection, we first generate a potentially unsafe raw command $u_\text{raw}\in\mathcal{U}$ and 
apply the filtered control $u_\text{filtered}$ by solving the following QP:
{\setlength\abovedisplayskip{4pt}
 \setlength\belowdisplayskip{4pt}
\begin{gather} \label{QP filter}
    u_\text{filtered} =  \arg\min_{u\in\mathcal{U}} \|u-u_\text{raw}\|_2^2 \\ 
    \text{s.t.} \quad 
    \underbrace{a^\lambda_\phi(x) u - \max_{u' \in \mathcal{U}} a^\lambda_\phi(x) u' + b^\lambda_\phi(x)}_{\partial v^\lambda_\phi(x,u)} 
    +\alpha(v^\lambda_\theta(x))\geq0 \nonumber
\end{gather}}
Here, $\alpha(\cdot)$ is a user-defined extended class-$\mathcal{K}$ function that regulates the aggressiveness of the filter.  
For simplicity, it is designed as $\alpha(z)=\alpha z$ with positive gain $\alpha >0$.
A larger $\alpha$ permits inputs that reduce the estimated safety value more rapidly. Before solving the QP, we solve a Linear Program (LP) to compute $\max_{u\in\mathcal{U}} a^\lambda_\phi(x)u$. When the networks are not yet sufficiently converged during training, the QP~\eqref{QP filter} can occasionally become infeasible. In such cases, we apply the control input that achieves the LP optimum instead. Since $a^\lambda_\phi(x) \approx \frac{\partial V^\lambda}{\partial x}(x)g(x)$, the LP fallback takes a greedy ascent step on the critic's estimated safety value. Finally, following \textbf{Corollary~\ref{corollary: discounting approach}}, 
the discount factor $\lambda$ is gradually decreased during training, so that the learned discounted value $v^\lambda_\theta$ converges to the undiscounted value function $V$.
\begin{algorithm}[t]
    \caption{Online Model-free Safety Filter Learning}
    \label{alg:safety_filter_learning}
    \KwIn{System with constraint signal, observation interval $\delta t$, raw input generator $\pi_\text{raw}(x,t)$, $\lambda_\text{init}>0$, $\lambda$-scheduler, mini-batch size $N$, learning rates $\gamma_\theta, \gamma_\phi$, target update ratio $\tau$ }
    \textbf{Initialize} Replay Buffer $\mathcal{B}=\{ \}$, Networks $\theta, \phi, \theta^\prime \leftarrow \theta, \phi^\prime \leftarrow \phi$, Discount factor $\lambda\leftarrow\lambda_\text{init}$\\ 
    \For{ episode $n=1, 2, \dots$}{
        Observe the state $x$ and the constraint $c$ from the system \\
        \For {t = $\delta t$, $2\delta t$, \dots}{
        Generate a raw input $u_\text{raw} \gets \pi_\text{raw}(x,t)$ 
        
        Infer the current safety value and derivatives $v^\lambda_\theta(x), a^\lambda_\phi(x), b^\lambda_\phi(x)$ 

        $u\gets$ Solve QP \eqref{QP filter} \textbf{if} QP is feasible \textbf{else}           $\arg\max_{u\in\mathcal{U}} a^\lambda_\phi(x)u$\;

        Apply $u$ to the system and observe transition $x^u_{\delta t}, c^u_{\delta t}$
        
        Store transition $\mathcal{B}\gets\mathcal{B}\cup(x, u, c, x^u_{\delta t}, c^u_{\delta t})$
        
        Sample a mini-batch of $N$ transitions from $\mathcal{B}$\;
        
        Compute loss $\mathcal{L}_v, \mathcal{L}_{\partial v}$ using \eqref{loss: V}, \eqref{loss: derivatives}
        
        Update network parameters $\theta \gets \theta-\gamma_\theta \cdot\nabla_\theta \mathcal{L}_v, \phi \gets \phi-\gamma_\phi \cdot\nabla_\phi \mathcal{L}_{\partial v}$

        Update target networks $\theta^\prime \gets(1-\tau)\cdot\theta^\prime+\tau\cdot\theta,\; \phi^\prime\gets(1-\tau)\cdot \phi^\prime+\tau\cdot\phi$ 

        Decrease $\lambda$ by the $\lambda$-scheduler
        
        $x\gets x^u_{\delta t}; \; c\gets c^u_{\delta t}$ \textbf{if} $c^u_{\delta t}\geq0$ \textbf{else} break    
        }
    }
\end{algorithm}
The proposed online model-free safety filter learning algorithm is summarized in \textbf{Algorithm~\ref{alg:safety_filter_learning}}. Although designed for online learning, it can also leverage offline data since the learning is off-policy.

\vspace{-5pt}
\section{Experiments}
Our method is validated through various dynamical systems, including Gymnasium environments \citep{towers2024gymnasium} popular in model-free RL\footnote{Source code: \href{https://github.com/snu-larr/Deep-QP-Safety-Filter}{https://github.com/snu-larr/Deep-QP-Safety-Filter}.}.
We highlight two key results: (1) consistent convergence to analytic solutions regardless of~$\delta t$, and (2) the least conservative behavior among baselines due to the reachability-based formulation.
We also validated the method on Hopper, a contact-driven hybrid system, confirming the practical relevance of the underlying assumptions. As a diagnostic, after training, we measured QP infeasibility over $10^6$ filtering calls: 0 occurrences on Double Integrator, Inverted Pendulum, and Inverted Double Pendulum, and 0.2343$\%$ on Hopper.

All QPs in~\eqref{QP filter} were solved using the \texttt{proxqp} solver~\citep{exp:ProxQP} with the \texttt{qp\allowbreak solvers} package~\citep{exp:qpsolvers}.
The discount factor $\lambda$ was initialized at $0.1/\delta t$ and gradually decayed to $0.0001/\delta t$, while the learning rate decreased from $3\times10^{-4}$ to $10^{-6}$.
Both schedules followed a polynomial decay with a power of 5 and decay intervals of $10^6$.
We constructed neural networks using ELU activations~\citep{elu_activation} and layer normalization \citep{ba2016layernorm}.
It consisted of two hidden layers with a width of 256, except for the Hopper environment, which used three layers.
We trained the models using the Adam optimizer~\citep{kingma2017adammethodstochasticoptimization} with $N=256$ and $\tau=0.005$. The raw input was generated by a random process $du_t=\kappa (\mu -u_t)dt+\sigma dW_t$, with $\kappa, \mu,$ and $\sigma$ randomly sampled from uniform distributions for every episode.
Since the safe set rendered by the safety value is scale-invariant\footnote{Multiplying the safety value by any positive constant does not change the safe set, i.e., $\mathcal{S}=\{x\;|\; V(x)\ge0\}=\{x\;|\; kV(x)\ge0\}\;\forall k>0$.}, we normalized each newly observed constraint value by its running maximum $c_{\max}$ to ensure that the constraint signal remained bounded by 1.
We also employed two value functions, $v^\lambda_{\theta,1}$ and $v^\lambda_{\theta,2}$ to balance over- and underestimation of the safety value.
When computing the next target value $v^\lambda_{\theta'}(x^u_{\delta t})$, we used the minimum of the two estimates, whereas the current target value $v^\lambda_{\theta'}(x)$ was computed using only one of them.
\vspace{-5pt}
\begin{figure}[t]
    \begin{minipage}[t]{0.6\textwidth}
        \centering
        \makebox[0.08\linewidth]{}
        \makebox[0.17\linewidth][c]{\small Analytic}
  \makebox[0.17\linewidth][c]{\small$\delta t = 0.1$} 
  \makebox[0.17\linewidth][c]{\small$\delta t = 0.05$} 
  \makebox[0.17\linewidth][c]{\small$\delta t = 0.01$} 
  \makebox[0.17\linewidth][c]{\small$\delta t = 0.005$} 
\includegraphics[width=\linewidth]{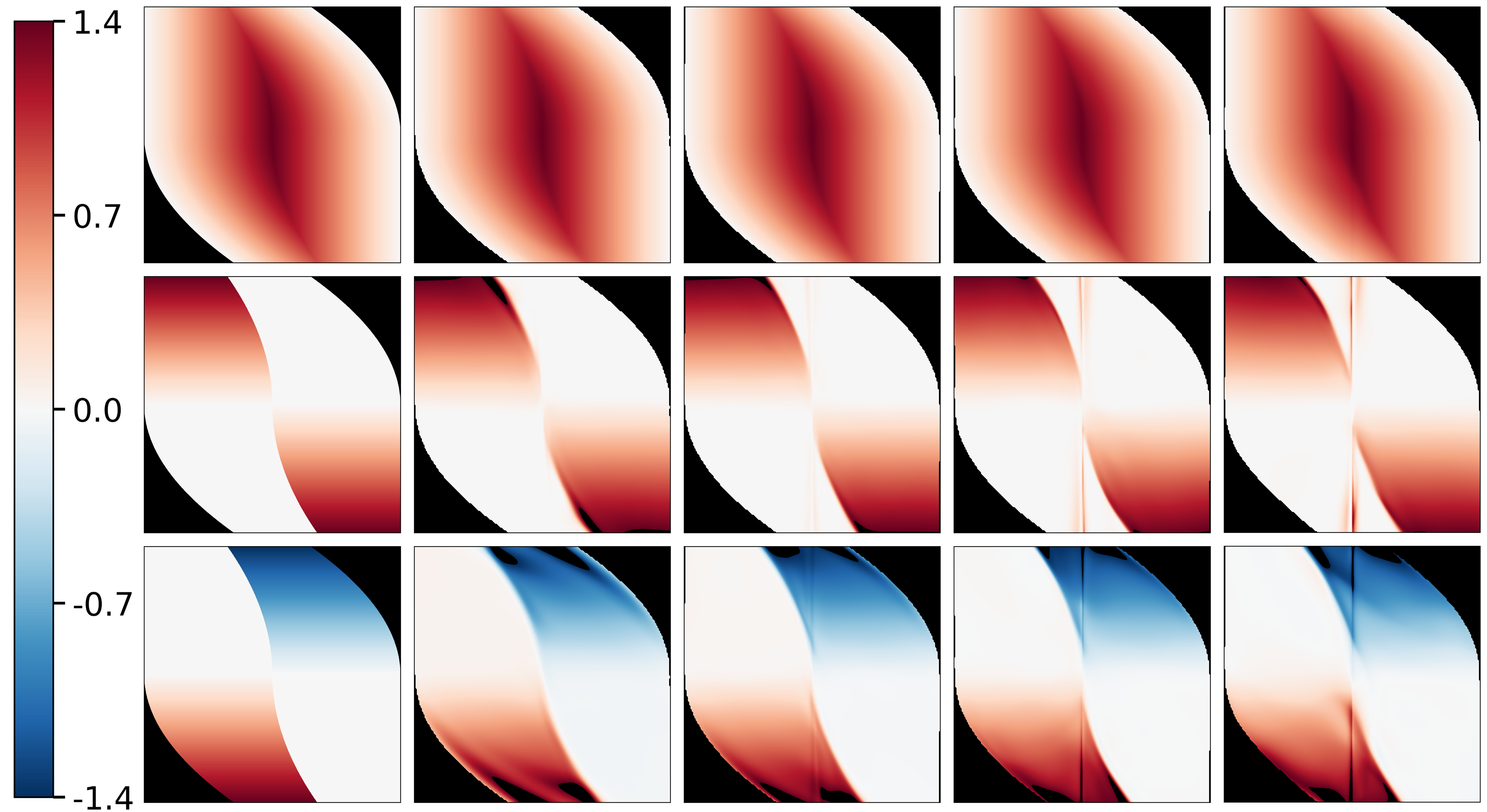}
  \caption{Comparison with analytic solutions. For each plot, the horizontal and vertical axes represent the position and velocity of the Double Integrator. From top to bottom, the first column shows the analytic solutions $V(x)$, $\max_{u\in\mathcal{U}}\partial V(x,u)$, and $\frac{\partial V}{\partial x}(x)g(x)$, while the remaining columns show the learned counterparts 
$v^\lambda_\theta(x)$, $b^\lambda_\phi(x)$, and $a^\lambda_\phi(x)$ under different~$\delta t$.}
  \label{fig:analytic_solution}
    \end{minipage}%
    \hfill
    \begin{minipage}[t]{0.37\textwidth}
        \centering
        \makebox[0.49\linewidth][c]{\small (a): $\alpha = 0.5$} 
        \makebox[0.49\linewidth][c]{\small (b): $\alpha = 5$} \includegraphics[width=0.92\linewidth]{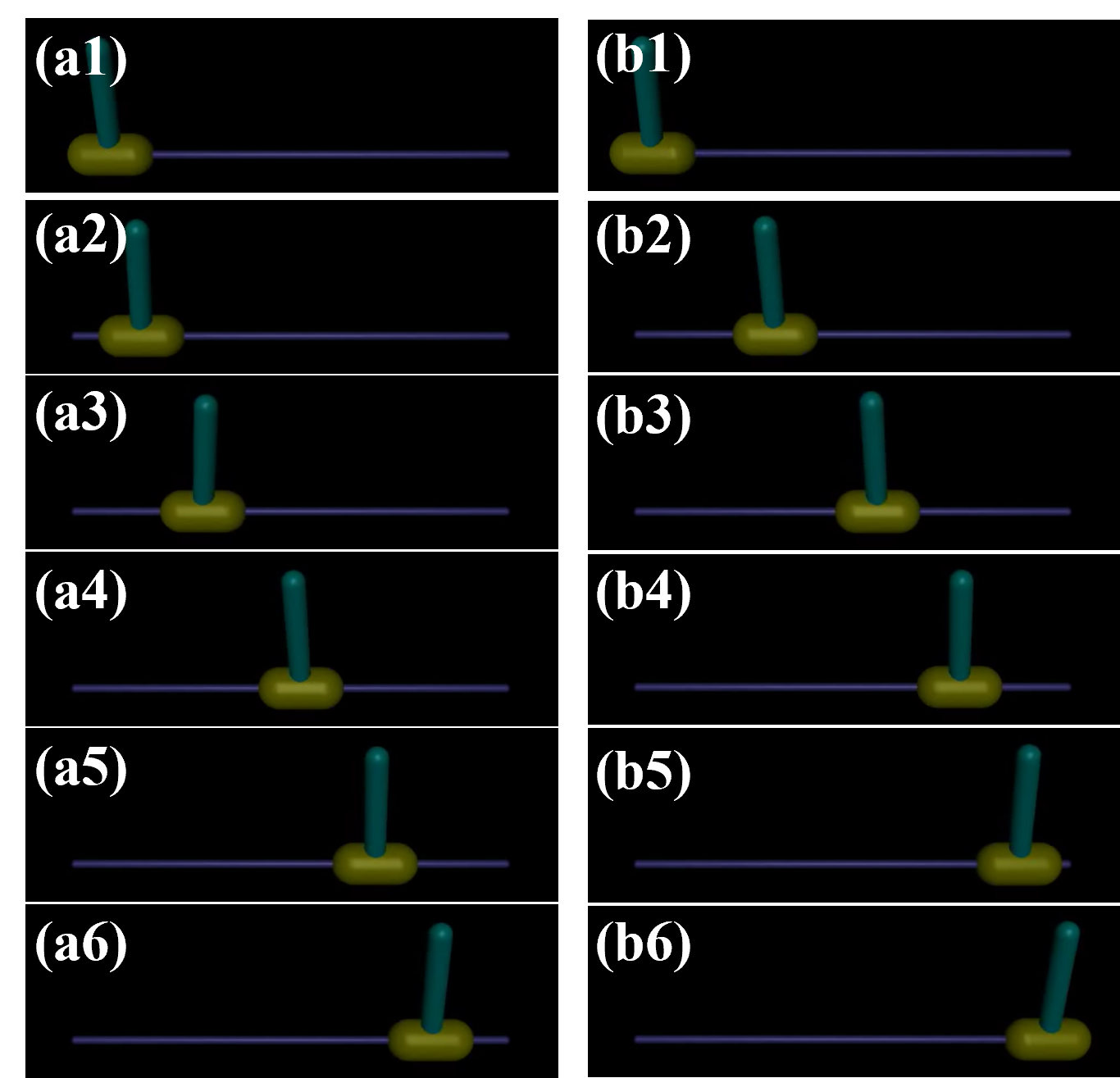}
        \caption{Effects of $\alpha$ on the aggressiveness. In both cases, $u_\text{raw}$ is switched from -1 to 1, and the two images in each row are captured at identical time steps. Larger~$\alpha$ produces more aggressive behavior with faster acceleration and deceleration.}
    \label{fig:alpha_difference}
    \end{minipage}
    \vspace{-15pt}
\end{figure}

\subsection{Convergence to the analytic solution}
We conducted experiments on the Double Integrator system. The constraint was set such that the position remains within the interval of $[-1.4, 1.4]$. We compare the learned critic against the analytic solution. The results are presented in Figure~\ref{fig:analytic_solution}, where only the safety domain of $V(x)\geq0$ or $v^\lambda_\theta(x)\geq0$ is plotted. The critic consistently converges across different $\delta t$ and captures the discontinuities in the analytic solution. \vspace{-5pt}

\subsection{Adjusting the aggressiveness via the free parameter $\alpha$}
The parameter $\alpha$ in \eqref{QP filter} is not learned but selected by the user to freely control the aggressiveness of the filter. Figure~\ref{fig:alpha_difference} shows the effect of different $\alpha$ values for the Inverted Pendulum environment with the constraint $c(x)=\min\{1-|x_\text{base}|,\,0.2-|\theta_\text{tip}|\}$. As $\alpha$ increases, the controller behaves more aggressively while preserving safety. In practice, 
$\alpha$ can be set considering the sustainable limits of the hardware. Furthermore, although the constraint is defined as the minimum of two functions with different scales, the learned filter maintains safe operation.

\begin{figure}[t]
    \centering
    \footnotesize
        \makebox[0.04\linewidth][c]{}
        \makebox[0.53\linewidth][c]{ Inverted Double Pendulum-v5}
  \makebox[0.29\linewidth][c]{Hopper-v5} 
    \includegraphics[width=0.9\linewidth]{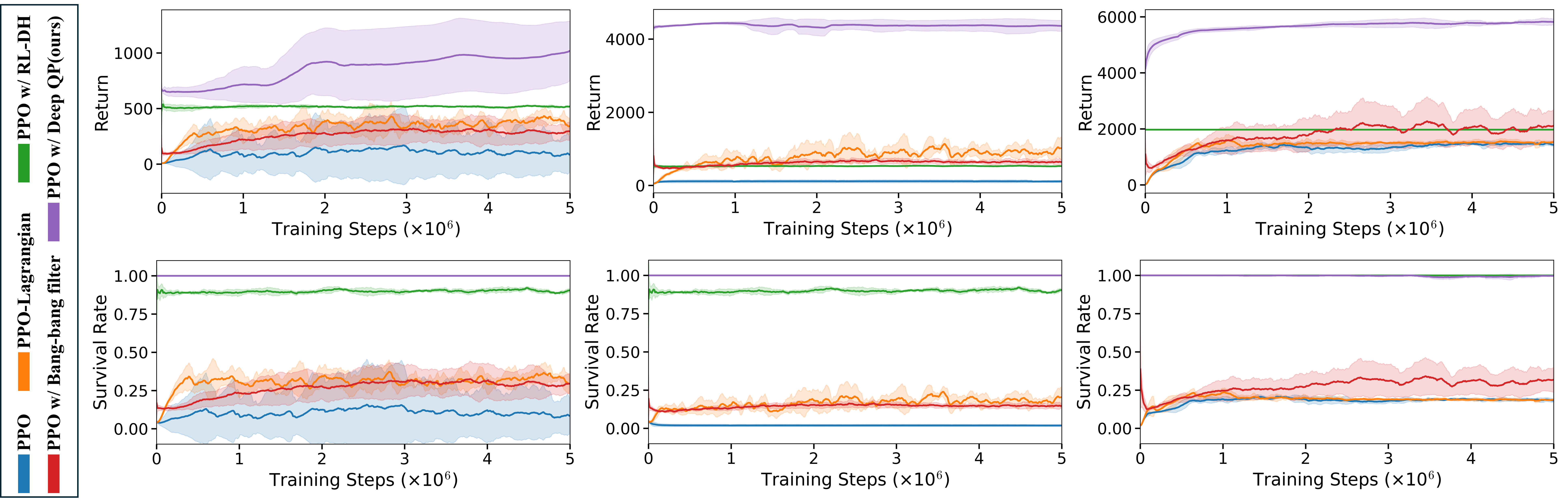}
        \makebox[0.06\linewidth][c]{}
        \makebox[0.27\linewidth][c]{(a)} 
        \makebox[0.29\linewidth][c]{(b)}
        \makebox[0.27\linewidth][c]{(c)} 
  \caption{Comparison Results of RL tasks with the learned safety filter against other baselines. Inverted Double Pendulum environment with modified rewards  (a): $|x|_\text{base}$ and (b): $|v|_\text{base}$. Hopper with default reward (c). All parameters of PPO used across baselines are identical.}
  \label{fig: fig4 RL}
    \vspace{-15pt}
\end{figure}

\subsection{Safe reinforcement learning}
Our safety filter is learned task-agnostically: its training objective depends only on the safety signal 
$c(x)$, not on any task reward or performance objective. Thus, the same filter can be paired with any RL controller.
Therefore, in this last subsection, we demonstrate the synergy between our learned safety filter and RL by showing that it can drastically reduce failures before convergence. The systems used are the Inverted Double Pendulum with $c(x)=\min\{0.95-|x_\text{base}|, y_\text{tip}-1\}$ and Hopper with $c(x)=\min\{z_\text{torso}-0.7, 0.2-|\theta_\text{torso}|\}$. Both systems are highly unstable and require sophisticated control strategies to satisfy the safety constraints. For the Inverted Double Pendulum, we considered two tasks with conflicting rewards:
 $|x_\text{base}|$ and $|v_\text{base}|$. The Hopper task inherently conflicts with safety because moving forward requires riskier motions. Baselines include PPO~\citep{schulman2017proximal}, PPO-Lagrangian~\citep{Ray2019_ppolangragian}, PPO with RL-DH~\citep{lavanakul2024_discriminatinghyperplane}, and PPO with our own variant implemented as a bang-bang safety filter $(\alpha\rightarrow\infty)$ with a switching criterion of $v^\lambda_\theta(x)<0.2$. Figure~\ref{fig: fig4 RL} shows that, eventhough the system is highly unstable and rewards are conflicting with the safety, the proposed method achieves faster and higher rewards with significantly fewer failures. The bang-bang variant further demonstrates that switching-based filters are unsuitable for such systems.
\vspace{-5pt}

\section{Conclusion}
We presented Deep QP Safety Filter, a fully data-driven, model-free safety layer for a black-box system that provides no prior information about its dynamics. By casting safety as a time-discounted reachability problem and proving contraction of the associated Bellman operators, our method stably learns both the safety value and its derivatives. In the exact setting, the critic converges to the viscosity solution, and empirically it remains robust even when the true value is non-smooth.
Experiments show that our filter preserves safety while being markedly less conservative than prior methods, achieving a practical balance between safety and task performance. We further demonstrate task-agnostic deployment: a single trained filter adapts across tasks via a scalar gain 
 that controls aggressiveness. When combined with RL, the filter substantially reduces failure rates during training without degrading returns, and continues to perform beyond idealized assumptions when realized with neural networks.
Overall, Deep QP Safety Filter delivers a principled, scalable safety mechanism for model-free control that is both theoretically grounded and practically effective.

\newpage
\acks{This work was supported by the National Research Foundation of Korea(NRF) grant funded by the Korea government(MSIT)(RS-2024-00436984).}

\bibliography{bjkim_ref}

\end{document}